\documentclass[a4paper,twoside]{article}

\usepackage[utf8]{inputenc}

\usepackage{epsfig}
\usepackage{subcaption}
\usepackage{calc}
\usepackage{amssymb}
\usepackage{amstext}
\usepackage{amsmath}
\usepackage{amsthm}
\usepackage{multicol}
\usepackage{pslatex}
\usepackage{apalike}
\usepackage{todonotes} 
\usepackage{hyperref} 
\usepackage{enumerate}
\usepackage{enumitem} 
\usepackage[autolinebreaks,useliterate]{mcode}
\usepackage{lipsum}

\usepackage{SCITEPRESS}  % Please add other packages that you may need BEFORE the SCITEPRESS.sty package.

\begin{document}

\title{cMinMax: A Fast Algorithm to Find the Corners of an N-dimensional Convex Polytope}

\author{\authorname{Dimitrios Chamzas\sup{1,2},
Constantinos Chamzas \sup{3}
and Konstantinos Moustakas\sup{1}} % and Third Author Name\sup{2}\orcidAuthor{0000-0000-0000-0000}}
\affiliation{\sup{1}Department of Electrical and Computer Engineering, University of Patras,  Rio Campus, Patras 26504, Greece}
\affiliation{\sup{2}McCormick School of Engineering,  Northwestern University, 2145 Sheridan Road, Evanston, IL 60208 USA}
\affiliation{\sup{3}Department of Computer Science, Rice University, Houston, TX 77251, USA}
\email{chamzas95@gmail.com, chamzask@gmail.com, moustakas@ece.upatras.gr }
}


%% Abstract section.
\abstract{During the last years, the emerging field of Augmented \& Virtual Reality (AR-VR) has seen tremendous growth. At the same time there is a trend to develop low cost high-quality AR systems where computing power is in demand. Feature points are extensively used in these real-time frame-rate and 3D applications, therefore efficient high-speed feature detectors are necessary. Corners are such special features and often are used as the first step in the marker alignment in Augmented Reality (AR). Corners are also used in image registration and recognition, tracking, SLAM, robot path finding, 2D or 3D object detection and retrieval as well as in linear programming algorithms.  Therefore there is a large number of corner detection algorithms but most of them are too computationally intensive for use in real-time applications of any complexity. Many times the border of the image is a convex polygon. For this special, but quite common case, we have developed a specific algorithm, cMinMax. The proposed algorithm is faster, approximately by a factor of 5 compared to the widely used Harris Corner Detection algorithm. In addition is highly  parallelizable. The algorithm is suitable for the fast registration of markers in augmented reality systems and in applications where a computationally efficient real time feature detector is necessary. The algorithm can also be extended to N-dimensional polyhedrons. %
} % end of abstract 

%%%%%%%%%%%%%%%%%%%%%%%%%%%%%%%%%%%%%%%%%%%%%%%%%%%%%%%%%%%%%%%%
%%%%%%%%%%%%%%%%%%%%%% START OF THE PAPER %%%%%%%%%%%%%%%%%%%%%%
%%%%%%%%%%%%%%%%%%%%%%%%%%%%%%%%%%%%%%%%%%%%%%%%%%%%%%%%%%%%%%%%%

\onecolumn \maketitle \normalsize \setcounter{footnote}{0} \vfill

\section{\uppercase{Introduction}}
\label{sec:introduction}

\begin{figure}[h!]
\centering
\includegraphics[width=0.95\linewidth]{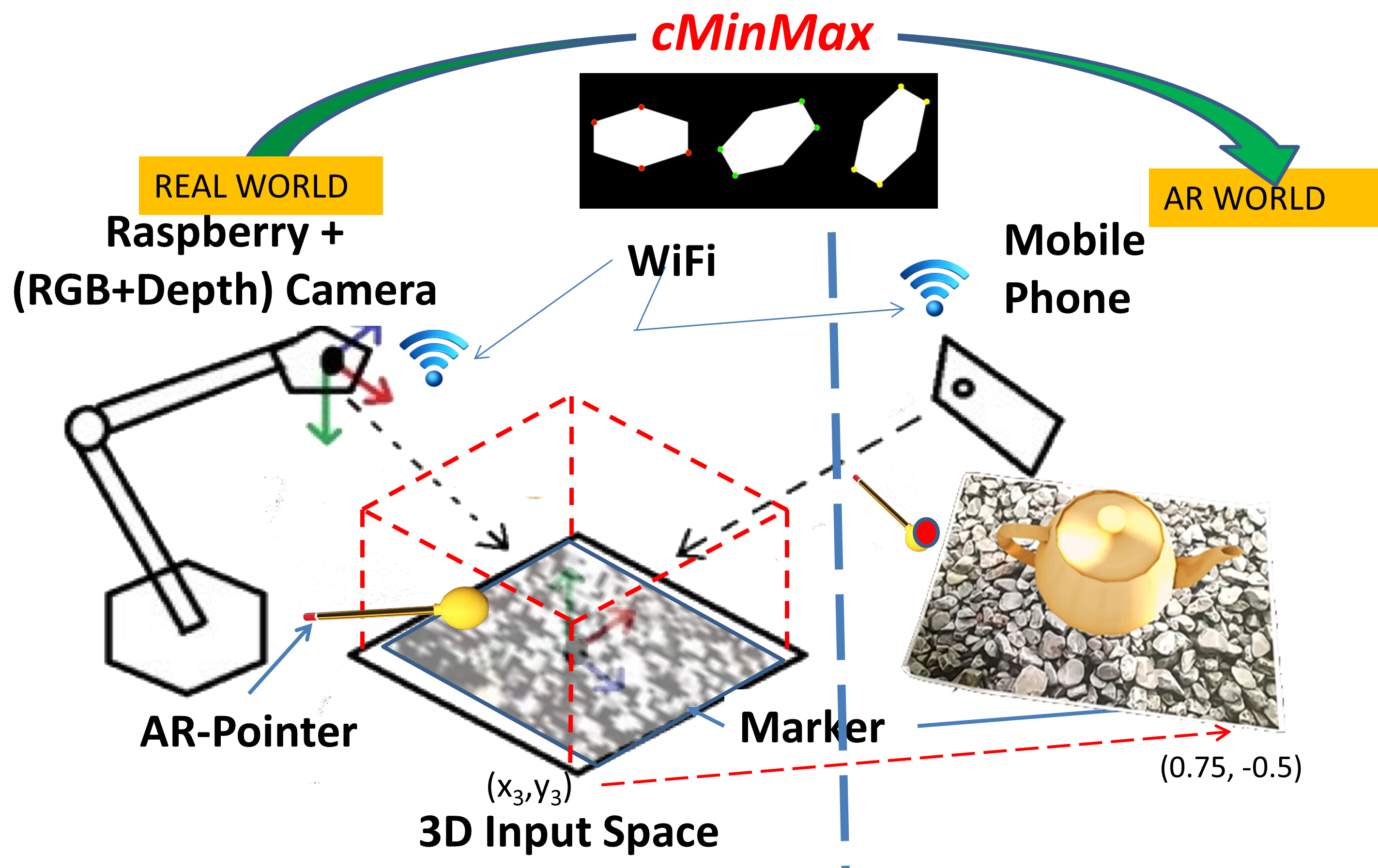}
 \caption{3D Augmented Reality Tangible User Interface using Commodity Hardware Using cMinMax to register Real World to AR World }
 \label{fig:systemLayout}
 \end{figure}

\noindent Augmented \& Virtual Reality (AR-VR) systems and applications have seen massive development and have been studied extensively over the last few decades \cite{billinghurst2015survey}. 
Also with the development of three-dimensional measuring technologies (3D Scanners) it is possible to acquire three-dimensional data using inexpensive three dimensional scanners raising the expectation that three-dimensional data and interfaces will be used. At the same time there is a trend to develop low cost high-quality 3D AR systems where computing power is in demand. \autoref{fig:systemLayout} shows such a low cost  3D Augmented Reality system using a tangible interface and constructed using commodity hardware \cite{chamzasD}.  Its central processing unit is a Raspberry Pi 4 equipped with a Raspberry camera.
Moreover smartphones are continuously evolving, adding more computer power, more sensors, and high-quality display. Multi cameras and depth sensors are some of their recent additions. Therefore, we expect that it will be possible to implement all the functionalities of an AR system just in a \textit{smartphone}.  In this cases, computing power will be in demand and we will need to develop new fast and efficient algorithms. One of the main problems in these systems is the registration of the Real and Virtual world, where we need to map the real-world 3D coordinates $(x_r, y_r, z_r)$ to the digital world coordinates $(x_v, y_v, z_v)$. One commonly used technique is the image marker.  We place an object, the marker, with a known shape in the real world and we want to find a projective transformation that will map this object to its virtual world counterpart. This transformation has to be recalculated every time the camera changes position within the real world environment and for real time systems this requires  a substantial amount of the systems computer resources. This becomes even worst when we are dealing with \textit{markerless} AR systems. A common approach to address this registration problem is finding features on the real world marker and since we know their position in the Virtual world, we can calculate the required projective transformation. Corners are such features.

 Detecting Corners is also the first step in many  Computer Vision and Object identification and retrieval tasks. It is also important to areas such as medicine, engineering, entertainment and so on  that are increasingly relying in processes that require this kind of information.
 In this work we present a simple and fast algorithm that addresses the above problem when the border of the image is a convex polygon.

\section{\uppercase{Previous Work}}
\noindent The problem to find the corners in an image was examined in the past. Most of the methods presented were based on the original algorithm proposed in \cite{harris1}, where they compute a corner by exploiting  sudden changes in image brightness. SUSAN \cite{smith1997susan} is another algorithm widely used for edge and corner detection.  
Using morphological operators was another approach \cite{lin1998modified} used to find the corners in an image.
A different approach  using machine learning was also  proposed in \cite{rosten2006machine}. 

With the development of three-dimensional technology and the usage of VR \& AR and Robotic systems,  another field that is growing fast over the last years is 3D or multidimensional data. Finding points of interest in 3D clouds \cite{nousias2020saliency,nousias2020mesh} or decomposing multidimensional workspaces into local primitives \cite{chamzasICRA}, becomes important and again corners (vertices) are one of them. An extension of Harris Corner Detection algorithm to 3D was proposed in \cite{glomb2009detection,sipiran2010robust,sipiran2011harris}. An example of extending SUSAN to 3D point clouds is described in \cite{walter2009susan} while in \cite{katsoulas2001efficient}  there is an indirect method that extracts edges from a 3D point cloud, and then regards these intersection
points as corners. In \cite{abe2017corner}, a technique is presented that estimates the vertices in a 3D Point Cloud on convex polyhedra surfaces using Delaunay Tetrahedralization. Convex Hull algorithms \cite{ComputationalGeometry2013,toth2017handbook} could also be used to determine the corners.

All of the above algorithms have a considerable processing cost as compared to the proposed technique, which is simple, robust and applicable to any dimension. Moreover is highly parallelizable. The input in the proposed method is a point cloud contained in a convex polytope acquired by an appropriate scanner. 

\section{\uppercase{The Algorithm}}
\noindent In image registration we often need to find the corners of the image. 
One of the most popular algorithms to address this problem is the Harris Corner Detection \cite{harris1,opencv03,opencv02} and its variants. 
Many times the border of the image is a convex polygon. For this special, but quite common case, we have developed a specific algorithm, referred as cMinMax. The algorithm utilizes the fact that if we find the x-coordinates of the pixels that belong to the image, then their maximum, $x_{max}$, is a corner's coordinate. Similarly for $x_{min}$, $y_{min}$ and $y_{max}$. The proposed algorithm is approximately 5 times faster than the Harris Corner Detection Algorithm, but its applicability is limited only to convex polygons.
%%%%%%%%%%%%%%%%%%%%%%%%%%%%%%%%%
\subsection{The Algorithm Steps for 2D}
The basic steps of the algorithm are:
\begin{enumerate}
    \item  Prepossessing: Generate a binary version of the image.
    \item If $\phi _ {max} = 2 \omega _ {max} $ is the expected maximum angle of the polygon, choose $\Delta \theta = \pi - \phi _ {max}$ and $M > \frac{  \pi} { 2(\pi - \phi _ {max})}=\frac{  \frac {\pi}{2}} { \Delta \theta}$ , 
    \item  For $k=0,1,..,M-1$, rotate the image by $ \theta _ {rotate} = k *  \Delta \theta = k ( \pi - \phi _ {max})$ 
    \item  Project the image on the vertical and horizontal axis and find the $( x_{min} , x_{max} , y_{min} , y_{max} )$. These are  coordinates of four corners of the rotated convex polygon.
   \item  Rotate the image backwards by $- \Delta \theta$  to the initial position and find the coordinates of the four corners.
    \item At the end, we have found $4 M$  points which is greater than the number of expected polygon corners. Hence, there are more than one pixels around each corner.
    We evaluate now the centroid for each of these bunches and these are the estimated corners of the convex polygon.
\end{enumerate}

Note: (a) The rotation step is always $\Delta \theta = \pi - \phi _ {max}$  but if we find only $x_{min}$ and $x_{max}$ then $M > \frac{  \pi} { \Delta \theta}$  and if we find only $x_{max}$ then $M > \frac{ 2 \pi} { \Delta \theta}$.   (b) For a canonical polygon with N corners where we find only $ x _ {max}$,  we have $\phi _ {max} =\frac {N*\pi -2 \pi}{N} $ and $ M > \frac{2 * \pi}{\Delta \theta}=N$.

 \begin{figure}[h!]
  \centering
  \includegraphics[width=\columnwidth]{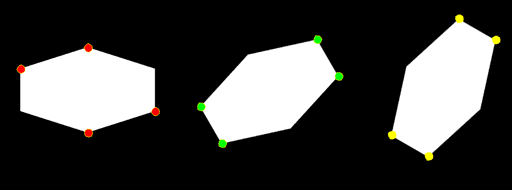}
  \caption{Detected corners in a hexagon for M=3 rotations. In each rotation we detect 4 corners}
  \label{fig:hexagon_rotate}
 \end{figure}
 
 In \autoref{fig:hexagon_rotate} we apply the algorithm in a hexagon. We have $\phi _ {max} = 120 ^ \circ$, thus $ M > \frac{  180 ^ \circ} { 2(180 ^ \circ - 120 ^ \circ)} = 1.5$, and we use M=3.

\subsection{The Proof}
\noindent
In this section we present the theoretical background for the algorithm.

\textit{The Problem:} Find the N-corners in an image that contains an object with a boundary that is a convex polygon.

\textit{Definition:}
Let us have a convex polygon with N-corners with coordinates $ I = ( x_i , y_i )_{i=1,N} $. One corner with coordinates $ ( x_k , y_k ) $ is called \textit{discoverable}, if one of its coordinates is maximum or minimum in the set $I$, that is 
% \vspace{-1cm}
\begin{align} \label{eq:minmax}
x_k &= max \ or \  min \ of \ (x_1 , x_2 , ... , x_N) \  or \\
y_k &= max \  or \  min \ of \ (y_1 , y_2 , ... , y_N) \nonumber
\end{align}

\begin{figure}[h!]
  \centering
  \includegraphics[width=0.80\columnwidth]{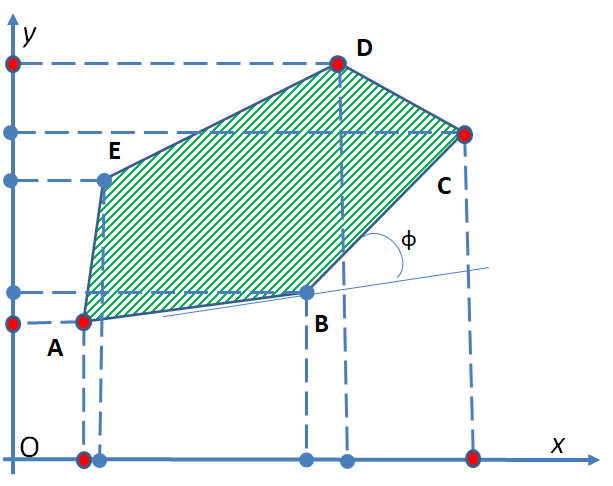}
  \caption{Discoverable Corners}
  \label{fig:pentagon}
 \end{figure}

\textit{Example:} In \autoref{fig:pentagon}, the corners $A,C,D$ of the pentagon $(ABCDE)$ are \textit{discoverable} , while $B,E$ are not.

\begin{figure}[h!]
  \centering
  \includegraphics[width=0.80\columnwidth]{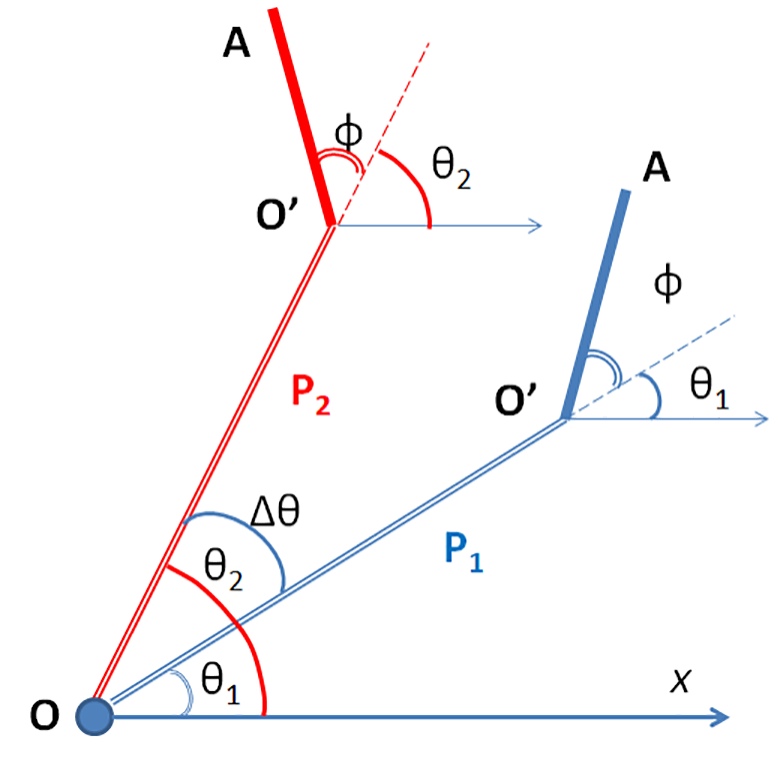}
  \caption{Rotating Rigidly Connected Lines}
  \label{fig:RotateCenter}
 \end{figure}

\textit{Proposition 1:} We have two connected lines $ (OO^{'})$ and $(O^{'}A)$ with the angle $\phi = \angle O O^{'} A $ to be constant (see \autoref{fig:RotateCenter}). If we rotate $(O O^{'})$ in increments of $ \Delta \theta$ around $O$, then $(O^{'}A)$  will rotate also in increments of $ \Delta \theta$ around  $O^{'}$.
 
\textit{Proof:}
Let us rotate $ (OO^{'})$ by $ \Delta \theta = \theta_{2} - \theta_{1}$, from position $P_1 $ to position $ P_{2} $. The  line segment $(O^{'}A)$  
rotates around $O^{'}$ from  $\theta_{1} + \phi $ to $ \theta_{2} + \phi $, and the change is again $ \Delta \theta = (\theta_{2} + \phi) - (\theta_{1}+\phi)$. Therefore if the line segment $ (OO^{'})$ rotates around $O^{'}$ in increments of $\Delta \theta $, then, neglecting translations, the line segment $(O^{'}A)$ rotates also in increments of $\Delta \theta $ around $O^{'}$ .

\textit{Proposition 2:} In  a convex polygon, $B$ is one of its corners, $\angle B$  is its angle and $\phi = 180 ^ {o} - \angle B $ its explementary. If we rotate the polygon around $B$ in increments of $\Delta \theta < \phi$, then in at most $M \geq \frac{2 \pi}{\Delta \theta}$ rotations, the adjacent points $A$ and $C$ will be at least once to the left side of vertical line $y y^{'}$.

 \begin{figure}[h!]
  \centering
  \includegraphics[width=0.80\columnwidth]{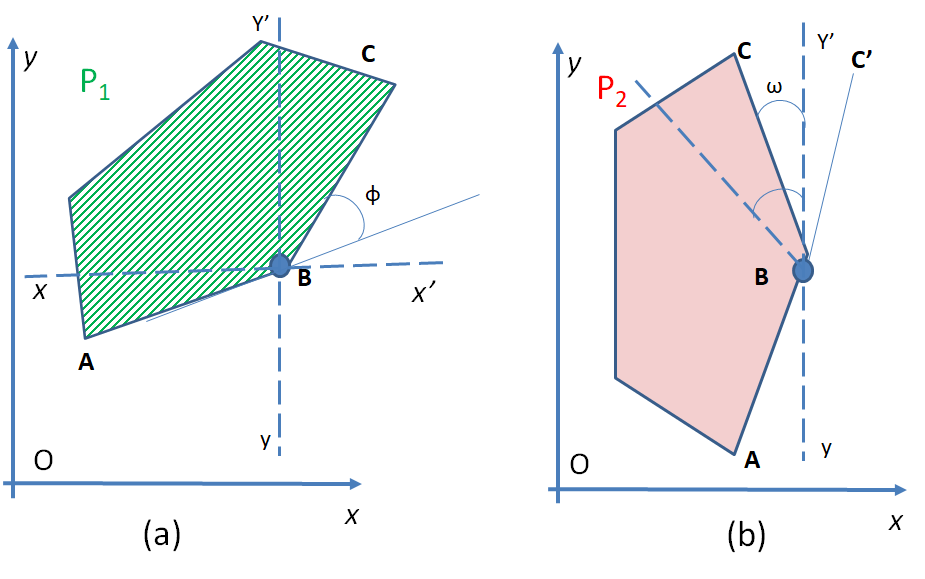}
  \caption{Invariant Rotation Center}
  \label{fig:RotatePolygone}
 \end{figure}
 
\textit{Proof:} Let us assume that we rotate the polygon counterclockwise around B, in increments of $\Delta \theta $ starting from position $P_{1}$  (Fig. \ref{fig:RotatePolygone} (a)). In $M$ steps, $ B C$ will make a full rotation around $B$.  Now let us consider position  $P_{2}$ when $ B C$ moves for the first time to the left of  $y y^{'}$ , (Fig. \ref{fig:RotatePolygone} (b)).
Then $\omega < \phi $   and $\omega + \angle CBA < \phi + \angle CBA = 180 ^ {o}$. Therefore point {A} is to the left of $y y^{'}$. Since the polygon is convex, all its corners are to the left of $y y^{'}$, therefore its coordinate $ x _{B} $, will be at least once the maximum of all the x-coordinates of the polygon angles.

\textit{Corollary 1:} For the corner $B$ to be $discoverable$, it is sufficient to rotate the polygon around $B$ in $M \geq \frac{pi}{2 \Delta \theta}$ with $ \Delta \theta < \phi$  (see Fig. \ref{fig:RotatePolygone}).

\textit{Proof:} In order for corner $B$ to be discoverable, it is enough for its two adjacent edges to be to the left of $y y^{'}$, or to the right of $y y^{'}$, or above $x x^{'}$ or below $x x^{'}$. Therefore we need only $M \geq \frac{pi}{2 \Delta \theta}$ rotation steps.

\textit{Theorem 1:} We have a convex polygon see Fig. \ref{fig:RotatePolygone} (a)), and let $\phi$ be the explementary of its maximum angle. We select a point $O$ and we rotate the polygon around it in increments of $\Delta \theta < \phi$, then in at most $M \geq \frac{ \pi}{2 \Delta \theta}$, all its corners will be \textit{discoverable} at least once.

\textit{Proof:} Let B a corner of the polygon. The angle between $OB$ and its adjacent edges is constant during the rotation around $O$. Then because of \textit{Proposition 1}, as we rotate the polygon around $O$ in increments of $\Delta \theta $, all its edges are rotating in increments of $\Delta \theta$ around their adjacent corners. Consequently according to \textit{Corollary 1} if we rotate the convex polygon around a point $O$ in increments of $\Delta \theta < \phi$, then in at most $M \leq \frac{ \pi}{2 \Delta \theta}$ steps, all its corners will be \textit{discoverable} at least once.

\subsection {Extension to N-dimensional Convex Polyhedrons}

\noindent The algorithm can also be extended to N-dimensional polyhedrons.

\textit{Definition 2:} A set C is \textit{convex} if for any points $x,y \in C$ the segment $[x,y]$ joining them belongs to C. A \textit{convex polyhedron} is a polyhedron that, as a solid, forms a convex set. Another definition is: A \textit{convex polyhedron} can also be defined as a bounded intersection of finitely many half-spaces \cite{Grunbaum1969Shephard,grunbaum2013convex}.

\textit{Definition 3:} 
Let us have a convex polyhedron with N-vertices with coordinates $ I = ( x_i , y_i, z_i )_{i=1,N} $. One vertex with coordinates $ ( x_k , y_k, z_k ) $ is called \textit{discoverable}, if one of its coordinates is maximum or minimum in the set $I$, that is 
% \vspace{-1cm}
\begin{align} \label{eq:minmax3D}
x_k &= max \ or \  min \ of \ (x_1 , x_2 , ... , x_N) \  or \nonumber \\
y_k &= max \  or \  min \ of \ (y_1 , y_2 , ... , y_N) \  or \\
z_k &= max \  or \  min \ of \ (z_1 , z_2 , ... , z_N) \nonumber
\end{align}  

\begin{figure}[h!]
  \centering
  \includegraphics[width=0.80\columnwidth]{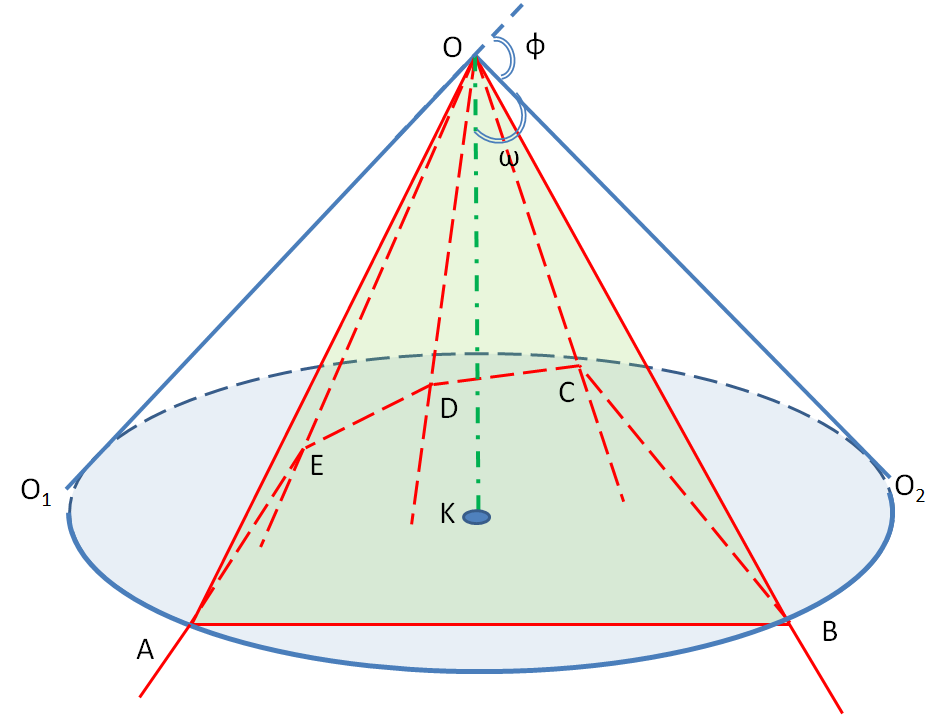} \caption{\textit{Minimum Bounding Cone} of a vertex}
  \label{fig:circumscribed_cone}
 \end{figure}

\textit{Definition 4:}  Let $O$ be a vertex of the convex polyhedron and $OA,OB,OC,OD,OE$ its edges (\autoref{fig:circumscribed_cone} (shown for 3D)). We define  $O_{1} O O_{2}$ as the \textit{Minimum Bounding Cone} for the vertex $O$, the smallest cone that its top is the vertex $O$ and it contains all the edges of the vertex $O$.   This \textit{Minimum Bounding Cone} will have at least two of the vertex edges on its surface and the rest inside. Let also $OK$ be its axis of symmetry. This way we can associate with each vertex of a convex polyhedron an angle, the angle $ \angle O_{1} O O_{2}=2 \omega $ of the \textit{Minimum Bounding Cone}. Since the polyhedron is convex, we have $ \phi > 0^{o} $ and $ 0^{o} < \omega = \angle KOO_{2} < 90^{o}$. The \textit{Minimum Bounding Cone Angle} is another way to define the solid angle of a vertex in an N-dimensional convex polytope  \cite{desario2011generalized}.

 \textit{Evaluation of the Minimum Bounding Cone angle:} Assuming that the vector $\overrightarrow{OA}$ is one of the edges that are on the surface of the circumscribed cone and the vector $\overrightarrow{OK}$ is its axis of symmetry, then the angle between these two vectors $ \angle AOK = \omega = \phi /2 $ i.e. $\omega = arccos \left( \frac {dot(\overrightarrow{OK}, \overrightarrow{OA})} {|\overrightarrow{OK}|| \overrightarrow{OA} |} \right )$. For a regular polytope, $A$ is any vertex adjacent to $O$ and $K$ is its center of gravity. 
 \begin{figure}[h!]
  \centering
  \includegraphics[width=0.99\columnwidth]{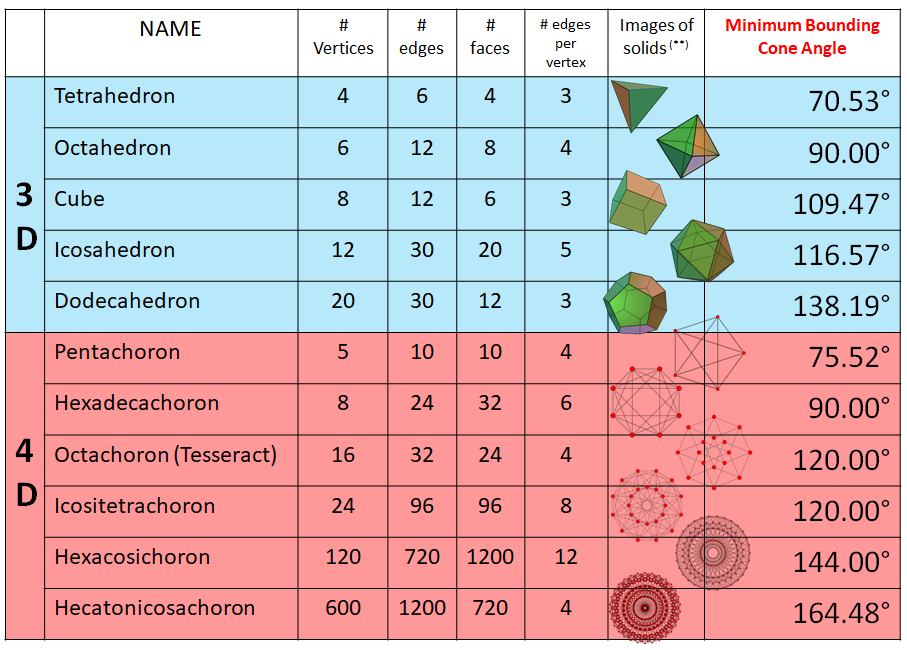} \caption{\textit{Minimum Bounding Cone Angles} for 3D and 4D Regular Polytopes}
  \label{fig:table_mbcAngle}
 \end{figure}
 In the last column of \autoref{fig:table_mbcAngle} we show the Minimum Bounding Cone Angles for the 3D and 4D regular polytopes.
 
\begin{figure}[h!]
  \centering
  \includegraphics[width=0.90\columnwidth]{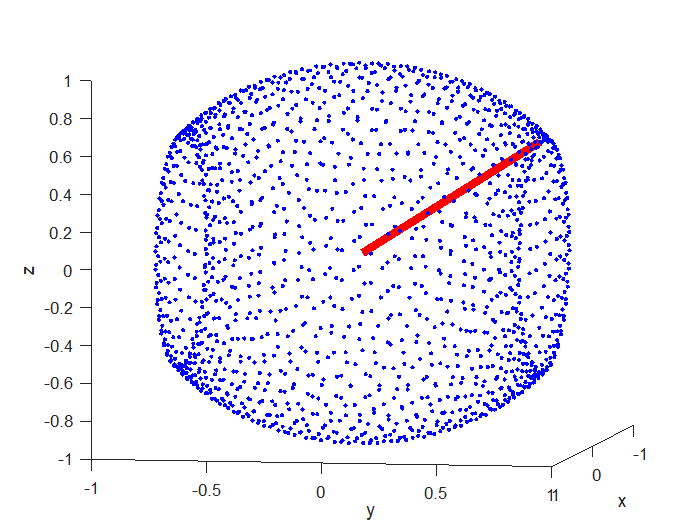} \caption{3D Grid of a vector position rotated around axis z and axis y with $\Delta \theta = \Delta \phi ={ \frac{\pi}{10}}$  rads }
  \label{fig:RotateGrid}
 \end{figure}

\textit{Proposition 3:} Let $R _ x ( \theta), R _ y ( \theta), R _ z ( \theta) $ be the rotation matrices by $\theta$ around axis $x,y,z$ respectively and
$ N _ { \theta } =\lceil {\frac{2 \pi}{\Delta \theta}} \rceil $ 
and $ N _ { \phi } = \lceil { \frac{2 \pi} {\Delta \phi}} \rceil$, where $\Delta \theta$ and  $\Delta \phi$ are incremental rotation angular steps around the axis $z$ and $z$.
We multiply a vector $\overrightarrow{OK }$ by the rotation matrix $ R _ {zy} = R _ y ( k \Delta \phi)  R _ z ( m \Delta \theta ) $ for $ k=0,1,2,..., N _ {\theta} -1$  and  $ m=0,1,2,..., N _ {\phi} - 1$. The $  N _ {\theta}  N _ {\phi}  $   positions of the rotated vector are shown in \autoref{fig:RotateGrid}.
At least for one of them, its distance $ d _ x $ from the axis $x$ is less than
$\frac{ \sqrt { { \Delta \theta} ^ 2 + {\Delta \phi} ^ 2 }}{2}$. The same is also true for $ d _ z $ the distance of a grid point from the axis $z$ .

\begin{figure}[h!]
  \centering
  \includegraphics[width=0.90\columnwidth]{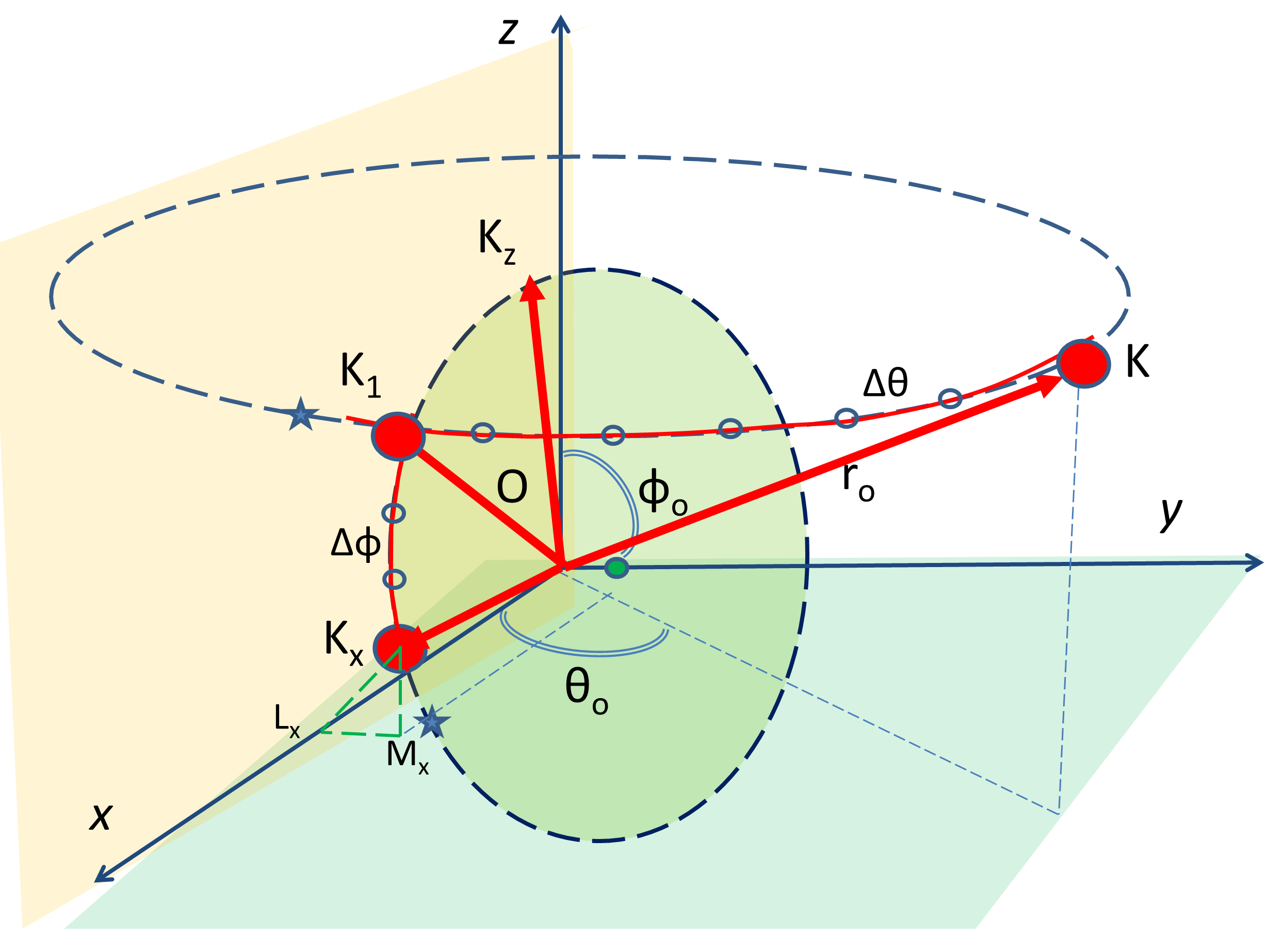} \caption{3D Rotation of a vector around axis z and axis y}
  \label{fig:RotateVectors}
 \end{figure}

\textit{Proof:}  We first rotate a unit vector $\overrightarrow{OK}$ around axis z by
$k \Delta \theta $ until it goes to its nearest position to plane (x,z).
This is position  $\overrightarrow{O  K  _ 1 }$ in \autoref{fig:RotateVectors}. Its distance from plane (xz) will be less than $\frac{\Delta \theta}{2}$.
Then we rotate it around axis y in steps of  $\Delta \phi$ until it goes to its nearest position to plane (x,y).  This is position  $\overrightarrow{O  K  _ f }$ in \autoref{fig:RotateVectors}.  Its distance from plane (xy) will be less than $\frac{\Delta \phi}{2}$.
Thus there is a pair  $( k _ {x} ,m _ {x})$ for which the vector $\overrightarrow{O  K  }$ goes to the grid position $\overrightarrow{O  K  _ x } $ and for this position its distance $d _x $ from the axis x is $d _x < {\frac{ \sqrt { { \Delta \theta} ^ 2 + {\Delta \phi} ^ 2 }}{2}}$.
Similarly for another pair $(k _ z , m _ z )$  the vector $\overrightarrow{OK}$  goes  to vector $\overrightarrow{O K _z  }$ the closest grid position to axis z. However, we can never go close to second axis of rotation,  axis y, since for any position in the grid its angle to axis y remains greater or equal to $ 90 ^ {o} - \phi$ (see \autoref{fig:RotateGrid}) \textsc{QED}. 

\textit{Proposition 4:} In  a convex polyhedron, the angle of the \textit{Minimum Bounding Cone} of vertex $O$ is $2 \omega$.  If we rotate the polyhedron first around axis z and then around axis y,  in increments of $k \Delta \theta $ and $ m \Delta \phi$, $k=0,...,N _ {\theta} - 1,m=0,2,...,N _ {\phi} - 1$ with $ \Delta \theta < \frac{\pi }{2} - \omega,  \Delta \phi < \frac{ \pi}{2} - \omega$ and  $N_{\theta} \geq \frac{\pi}{\Delta \theta}, N_{\phi} \geq \frac{ \pi}{\Delta \phi}$,  then the \textit{Minimum Bounding Cone} of vertex $O$ will fall at least once in the upper side of the plane vertical to axis z, passing from $O$ (see \autoref{fig:RotatePolygone}). Also similarly will be once bellow and once above the plane passing from $O$ and vertical to axis x. 
 
\begin{figure}[h!]
  \centering
  \includegraphics[width=0.90\columnwidth]{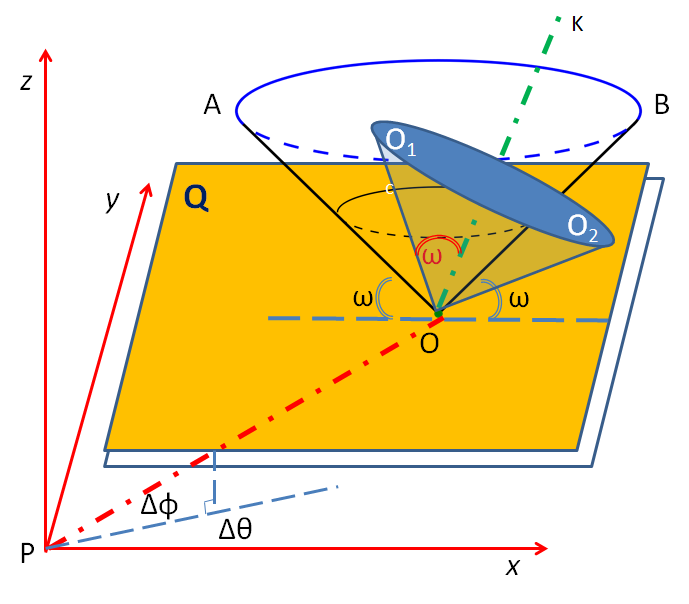} \caption{3D Rotation of a Convex Polyhedron}
  \label{fig:RotateVertex}
 \end{figure}

\textit{Proof:}  Similar to \textit{Proposition 2}. As we multiply the points of the polyhedron with the rotation matrix $R _zy$ the direction of vector $\overrightarrow{O K  }$ will go through all its corresponding position in its 3D grid.  As proved in \textit{Proposition 3},  at least one position of the 3D grid corresponding to the axis of symmetry $OK$ of the \textit{Minimum Bounding Cone} of vertex $O$ will fall  inside the cone $(OAB)$. This cone is perpendicular to the plane \textbf{Q} and the angle of its sides to $Q$ is $\omega$
 
\textit{Theorem 2:} We have a convex polyhedron, and let $\phi _ {max} = 2 \omega _ {max} $ be the \textit{maximum} angle of all the \textit{Minimum Bounding Cones} corresponding to its vertices. We select two axis for the 3D case (N-1 axis for an N-dimensional object) of the coordinate system. i.e. \textit{z} and \textit{y} and by multiplying all the points with the rotation matrix $R _ {zy} $, we rotate the polygon around them with  $k \Delta \theta$ and $ m \Delta \phi$ where $k=1,...,N_{\theta}, m=1,2,...,N_{\phi}$, $ \Delta \theta < { \frac{ \pi}{2} - \omega _ {max}}$, $ \Delta \phi < {\frac{ \pi}{2}  - \omega _{max}}$ and $ N _ {\theta} \geq \frac{ \pi}{ \Delta \theta},  N_{\phi} \geq \frac{  \pi}{ \Delta \phi}$. Then all its vertices will be \textit{discoverable} at least once in the axis x and axis z.   

\textit{Proof:} It follows from \textit{Proposition 4}.

\textit{Note:} A different line of proof could be based on the fact that the projection of convex polyhedron on a plane is a convex polygon.  Thus we rotate the polyhedron around an axis, we project it on the planes that contain the axis and then we apply the 2D algorithm to the obtained convex polygons.

The extension of the algorithm to N-dimensional Convex polytopes is possible in a similar way. Rotations must now defined in N-dimensional space \cite{aguilera2004general} around their "N-dimensional axis".

\section{\uppercase{Implementation}}

\noindent  A crucial parameter in the above algorithm was the choice of M. With $\phi _ {max}$ the expected maximum angle of the polygon, it was shown that if $M \geq {  \pi} / ({ \pi - \phi _ {max}})$, then each corner of the polygon will appear at least once in the set of the detected corners. For example, for an orthogonal parallelogram,  $\phi _ {max} =2 \pi / 4$ , $ M \geq 2$ and if we chose $ M=2$,  the rotation step is $\pi /4$. For a hexagon we have $\phi_{max} =2 \pi / 3$, $ M \geq 3$ and if we chose $ M=3$  the rotation step is $\pi /6$. However, when an edge becomes nearly vertical to an axis, due to numerical accuracy and noisy data, many times there are more than one $max$ or $min$ points in the projection on one axis. In this case we decided to neglect all of them and go the next rotation step. Thus, we must make more rotation steps than the one predicted by the theoretical analysis. Another parameter is the center of the image rotation. Again, as it was shown, we can choose any point as the image rotation center, but it is expected that if the rotation center is the centroid of the convex polygon, the algorithm to be less sensitive to numerical errors. 

\subsection{Examples}

\subsubsection {2D Case}
\begin{figure}[h!]
  \centering
  \includegraphics[width=0.90\columnwidth]{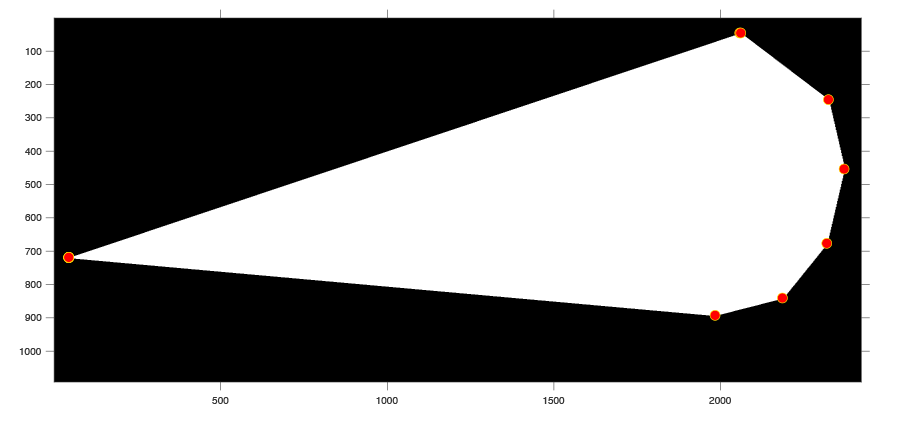} \caption{Estimated Corners in a heptagon (red dots).} 
  \label{fig:Eptagon}
 \end{figure}

\noindent For the 2D case, we used a 2040x1080  binary image of a  convex polygon with seven corners \autoref{fig:Eptagon} and $ {\phi _ {max}} \approx 158 ^ \circ$. The required number of rotations must be at least $M \geq \frac {180 ^ \circ} {180 ^ \circ - 158 ^\circ} = 8.53$. Thus we used N=9 rotations with $\Delta \theta = \frac {90 ^ \circ} {9} = 10 ^\circ$

\subsubsection {3D Case}
\begin{figure}[h!]
  \centering
  \includegraphics[width=0.90\columnwidth]{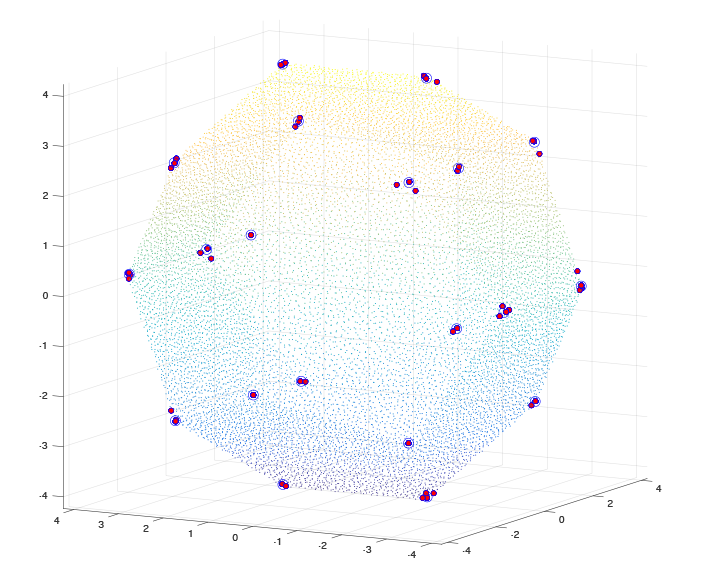} \caption{Estimated Corners in a dodecahedron (red dots). Circles indicate the weighted position of corners}
  \label{fig:dodecahedron}
 \end{figure}

\noindent  In this example we used a dodecahedron point cloud obtained from MeshLab, with 14535 points. The length of its edge is 3.2361 . 
The results for this dodecahedron  with $\Delta \theta = \Delta \phi = \frac {\pi}{20} $ are shown in \autoref{fig:dodecahedron}. We did 20x20=400 rotations, for every rotation we find 6 corners, 2 in each axis, but only  282 of them were accepted as valid and they were classified as corners. For the other cases, due to numerical accuracy we had more than one max or min in one axis and they were rejected.  These 282 points were clustered to 20 groups, and their centroids were the estimated corners. The average accuracy of the estimation was approximate 2\% of the edge length. The maximum angle of the minimum bound cone is $\omega _{max} = 69.095 ^ \circ$, and theoretically we could use $\Delta \theta = \Delta \phi = \frac {\pi}{9} $, but due to numerical errors and noise we need less than half of it.

\subsection { Evaluation: Computational Complexity}

\noindent Let us assume we have a point cloud with $n$ points of a N-dimensional polytope and we know that the  $\phi _ {max} = 2 \omega _ {max} $ is the expected \textit{maximum} angle of all the \textit{Minimum Bounding Cones} corresponding to its vertices. Then, following Theorem 2, with $ \Delta \theta < { \frac{ \pi}{2} - \omega _ {max}}$ and $ N _ {\theta} \geq \frac{ \pi}{ \Delta \theta}$, we have to perform $(N-1) N _ {\theta} ^ {(N-1)}$ rotations of $n$ points, in order all its vertices to be \textit{discoverable} at least once in one axis. Therefore in every step of the algorithm we perform one rotation and then find the $max$ of the n points x-coordinates. Both operations are of complexity $O(n)$ and we have to perform at least $L = {(N-1) N _ {\theta} ^ {(N-1)}}$ steps, thus the algorithm computational complexity is $LO(n)$. At this point we have to observe that each step is independent from the others, therefore they can computed in parallel and the proposed algorithm is highly parallelizable. Assuming that the algorithm is running in a computer with at least L GPUs then we can claim that its complexity is $O(n)$. Convex Hull and Harris corner Detection algorithms can also be used to address similar problems. Convex Hull algorithms are difficult to be parallelizable  and their sequential version is of $O(nlog(n))$ complexity  \cite{ComputationalGeometry2013,toth2017handbook}.
\begin{figure}[h!]
  \centering
  \includegraphics[width=0.90\columnwidth]{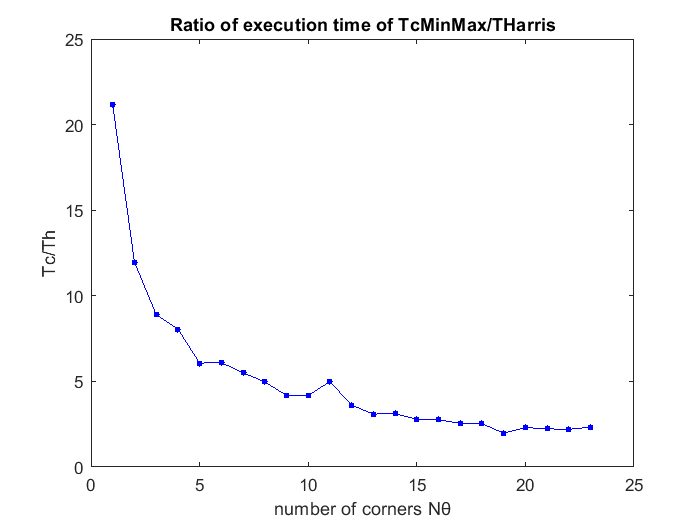} \caption{Ratio of execution time for cMinMax and Harris Corner Detection algorithm applied to regular polygons with 3 to 25 corners }
  \label{fig:timeratio}
 \end{figure}
To compare it with the complexity of Harris corner detection algorithm in 2D \cite{Chen2009}, we did run both of them in MatLab\textsuperscript{\textregistered}, using the $detectHarrisFeatures()$ command. For 2D space we have $ N =2$ and the complexity of $cMinMax$ is $ N _ {\theta}O(n)$. For images with 3-12 corners, cMiniMax is on the average 5 times faster than Harris Corner detection algorithm (see \autoref{fig:timeratio}). In addition the proposed algorithm appears to be less sensitive to sampling quantization errors.

\section {\uppercase{Random Sampling}}

\noindent Most of the times the number of unknown corners is not given and  in addition we do not have a good estimation of $\phi _ {max}$. Thus we cannot estimate a proper rotation step for the application of cMinMax. One approach will be to start with an initial rotation step. Next we reduce it and try again, until the number of detected corners remain constant. An alternative approach is to rotate the polytopes  with angles selected \textit{randomly},  In this case it is important to have uniformly distributed rotations. The 2D case is simple, but we have to be careful when we deal with objects in with dimensionality higher than two. 

\textbf{2-D}: We select a random angle $\Delta \theta$ in the closed interval $[-\pi, \pi]$. We rotate the convex polygon by $\Delta \theta$ and we find the extremes of the coordinates in the x-axis and y-axis. We continue until no more different corners are detected. 
\begin{figure}[h!]
  \centering
  \includegraphics[width=0.95\columnwidth]{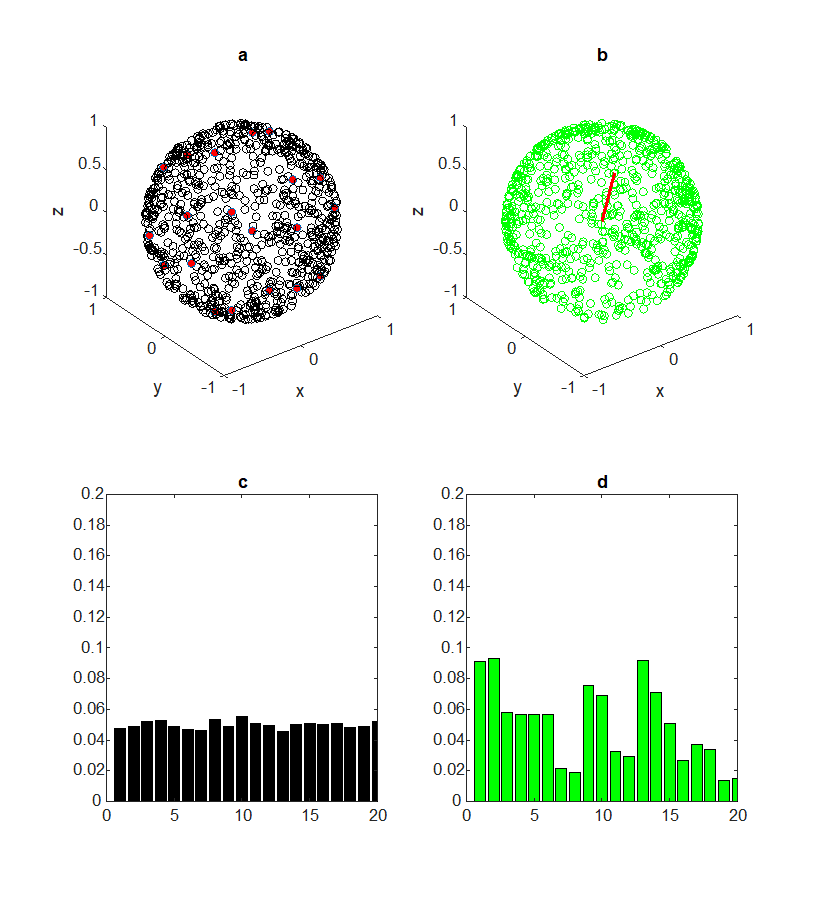} \caption{(a) Uniformly distributed points on a sphere. Red dots the 20 dodecahedron vertices. (b) A vector and its rotated positions. (c) Histogram of (a) around the 20 red dots. (d) Histogram of (b) around the 20 red dots.}
  \label{fig:histogramSphere}
 \end{figure}
 
\textbf{3-D}: We select two random angles $\Delta \theta$ and $\Delta \phi$. $\Delta \theta$ is uniformly distributed in the interval $[- \pi /2, \pi/2]$.  $\Delta \phi$ is randomly distributed in the interval $[-\pi, \pi]$ with a \textit{density distribution} $ f( \phi ) = sin ( \phi )/ 2$ . This way we have more points around the equator $\phi = 0 $, generating thus uniformly distributed pairs $\Delta \theta, \Delta \phi$ on a sphere \footnote{http://corysimon.github.io/articles/uniformdistn-on-sphere/}  (see \autoref{fig:histogramSphere} (a)). We rotate now the convex polyhedron by $\Delta \theta$ and $\Delta \phi$ and we find the extremes of the coordinates in the x-axis, y-axis and z-axis. 
To make the final position of the rotated points as random as possible, in every step we peak randomly one of the possible six possible axis rotations. The rotated position of a vector are shown in \autoref{fig:histogramSphere} (a). In \autoref{fig:histogramSphere} (c) and (d) we show the histograms for (a) and (b). Each of the 20 bins contain the points that close to the corresponding vertex of dodecahedron  (red dots in (a)). 

To simplify our analysis we will examine the case where we find the max ONLY in the x-axis. It is clear that in every rotation we detect only ONE corner. The question we want to answer is, how many times do we have to rotate a polytope with \textit{N} corners, in order to detect all its corners. This problem is equivalent to the following Die problem \cite{Isaac95}, irrespective of the dimensionality of the polytope space, \textit{ "Roll a die with N-faces.  What is the expected number of rolls to get all its N faces?}. \footnote{\url{http://www.cis.jhu.edu/~xye/papers_and_ppts/ppts/SolutionsToFourProblemsOfRollingADie.pdf}} 

\begin{figure}[h!]
  \centering
  \includegraphics[width=1\columnwidth]{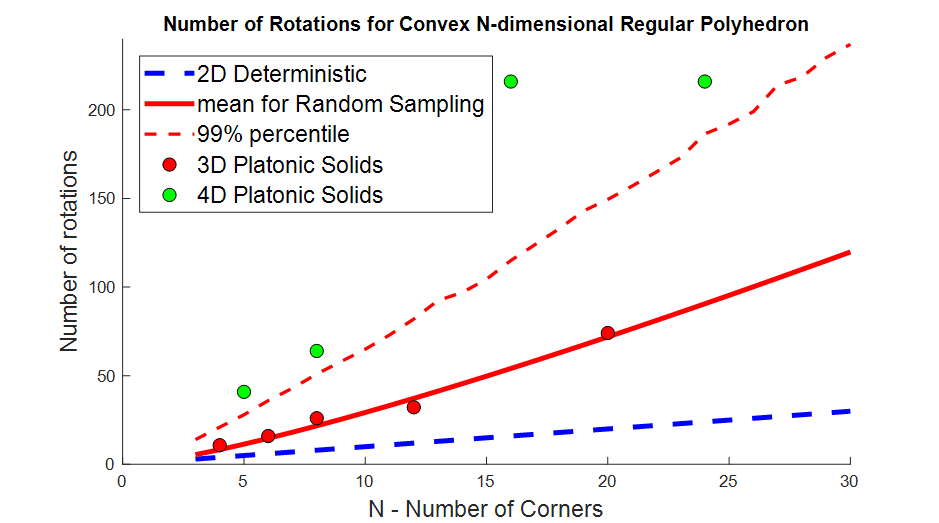} \caption{Mean and 99\% percentile of required Rotations for a canonical polytope with N corners (red line). With blue dashed line and with red and green circles are the required rotations for the deterministic case for 2D, 3D and 4D canonical polyhedrons}
  \label{fig:RandDeterm}
 \end{figure}
 
In \autoref{fig:RandDeterm}, we show the theoretically required number of rotations for 2D, 3D and 4D canonical polytopes with N vertices \cite{coxeter1973regular,paffenholz2017polydb}. For the case of rotating in equal angle steps, for the 2D canonical polygons we have $ N _ {rot} = \frac {2\pi}{\pi -2 \omega _ {max}} = N  $. For 3D and 4D polytopes we have $ N _ {rot} = {\left (\frac {2\pi}{\pi -2 \omega _ {max}} \right)} ^ 2 $ and $ N _ {rot} = {\left (\frac {2\pi}{\pi -2 \omega _ {max}} \right)} ^ 3 $ respectively. From \autoref{fig:table_mbcAngle}  we have for the 3D platonic solids with $N= 4,6,8,12,20$ vertices,  that their angles $\phi _ {max}= 2 \omega _ {max}$ of their Minimum Bound Cones are $70.53 {^\circ}, 90.00 {^\circ}, 109.47 {^\circ}, 116.57 {^\circ}$ and $ 138.19 {^\circ}$ respectively. There are six 4D convex platonic solids \footnote{\url{http://eusebeia.dyndns.org/4d/regular}},\footnote{\url{https://en.wikipedia.org/wiki/Regular_4-polytope\#Regular_convex_4-polytopes}} with $N= 5, 8, 16, 24,120,600$ vertices.  The angles $\phi _ {max}= 2 \omega _ {max}$ of their Minimum Bound Cones is for $N=5,16 $, $75.52 {^\circ}, 120 {^\circ},$ respectively. For the random case, a canonical polytope is equivalent to a \textit{fair} die with N equiprobable faces and it is known that the expected number of rolls to get all its N faces is the harmonic mean of N, i.e. $m _{N _ {rot}} = N \sum_{n=1}^{N}{\frac{1}{n}}$. This is the red line in \autoref{fig:RandDeterm} together with the 99\% percentile.
From \autoref{fig:RandDeterm} we conclude that for 2D it is preferable to use the rotation in equal steps, for 3D, rotation in equal steps and random rotation are  equivalent but for higher dimensions the random rotation case is expected to be preferable.

\section{\uppercase{Conclusions}}
\label{sec:conclusion}

\noindent  A new corner estimation technique on N-dimensional point clouds of convex polytopes was proposed in this contribution. The proposed algorithm is based on the fact that the min and max of projected coordinates in any axes belong to a corner. For 2D we compared it with the Harris corner detection algorithm (implementation at Matlab) and it was approximately 5 times faster for objects with less then 10 corners. We defined the solid angle of a vertex of an N-dimensional convex polyhedron by introducing the concept of the $Minimum Bounding Cone$ and we proved that the algorithm terminates in finite steps and the number of steps depends on the maximum solid angle of the convex polytope. We study 2 different techniques for rotating the point cloud of the object, either rotating by incremental angle steps(deterministic) or by choosing the angles of rotation randomly. We concluded that for 2D if preferable to use deterministic approach, for 3D the two methods are equivalent but for higher dimension we expect the random to be preferable. Another advantage of the algorithm is that it can be implemented using parallel processing since all the rotations can be executed simultaneously. A limitation of the proposed algorithm is that it requires a prior estimation of the maximum solid angle of the convex polytope and in future work we will try to address this problem. Usage in real time multidimensional applications is another one, so we plan to develop a faster version of the algorithm suitable for graphics cards using multiple GPUs by exploiting its parallel implementation. Finally, using morphological operators we will try to extend its applicability to non-convex objects,

\section*{\uppercase{Acknowledgments}}

\noindent
The authors wish to thank the members of the Visualization \& Virtual Reality Group of the Department of Electrical and Computer Engineering of the University of Patras as well as Dr. A. Koutsoudis and Dr. G. Ioannakis from the Multimedia Research Lab of the Xanthi's Division of the "Athena" Research and Innovation Center, for their useful comments and  discussions during the initial preparation of this work.
Constantinos Chamzas for this work was supported by the National Science Foundation, Graduate Research Fellowship Program under Grand NSF-GRFP 1842494 and Konstantinos Moustakas by the European Union’s Horizon 2020 research and innovation programme under Grant Agreement No 871738 - CPSoSaware-Crosslayer cognitive optimization tools \& methods for the
lifecycle support of dependable CPSoS.

\bibliographystyle{apalike}
{\small
 \bibliography{ChamzasBibliography}
}

\end{document}